\documentclass[runningheads]{llncs}
\def\shownotes{0}

\usepackage[usenames,dvipsnames]{xcolor}
\usepackage{subfig}

\usepackage[normalem]{ulem}
\usepackage{setspace}
\usepackage{times}
\usepackage{epsfig}
\usepackage{amsmath}
\usepackage{amssymb}
\usepackage{verbatim}
\usepackage{balance}
\usepackage{color}
\usepackage{xspace}
\usepackage{enumerate}
\usepackage{graphicx}
\usepackage[noend]{algpseudocode}
\usepackage{algorithm}
\usepackage{wrapfig}
\usepackage{microtype}
\usepackage{pifont}
\usepackage{caption}
\usepackage{multirow}
\usepackage{fancyhdr}
\usepackage[hyphens]{url}
\usepackage{hyperref}
\usepackage[margin=1.4in]{geometry}

\graphicspath{ {images/} }

\ifnum\shownotes=1
\newcommand{\authnote}[2]{{\textcolor{red}{\textsf{#1 notes: }\textcolor{blue}{ #2}}\marginpar{\textcolor{red}{\textbf{!!!!!}}}}}
\else
\newcommand{\authnote}[2]{}
\fi

\newcommand{\comments}[1]{\ignorespaces}

\usepackage[utf8]{inputenc} 
\usepackage[T1]{fontenc} 

\newcommand{\ignore}[1]{}

\newcommand{\Tspc}{\rule{0pt}{2.2ex}}
\newcommand{\TspcBig}{\rule{0pt}{2.5ex}}

\begin{document}

	\title{Unconstrained Iris Segmentation using Convolutional Neural Networks} 

	
\author{Sohaib Ahmad \and
	Benjamin Fuller}
\authorrunning{S.Ahmad, B.Fuller} 
\institute{University of Connecticut, Storrs CT 06269, USA 
	\url{www.cse.uconn.edu}
	\email{\{sohaib.ahmad,benjamin.fuller\}@uconn.edu}}
	
	\date{}
	\maketitle
	
	\thispagestyle{empty}
	\pagestyle{plain}

	\begin{abstract}
		The extraction of consistent and identifiable features from an image of the human iris is known as iris recognition.  Identifying which pixels belong to the iris, known as segmentation, is the first stage of iris recognition.  Errors in segmentation propagate to later stages. Current segmentation approaches are tuned to \textit{specific environments}. 
			
		We propose using a convolution neural network for iris segmentation.  \textit{Our algorithm is accurate when trained in a single environment and tested in multiple environments.} Our network builds on the Mask R-CNN framework (He et al., ICCV 2017).  Our approach segments faster than previous approaches including the Mask R-CNN network. 

		Our network is accurate when trained on a single environment and tested with a different sensors (either visible light or near-infrared).  Its accuracy degrades when trained with a visible light sensor and tested with a near-infrared sensor (and vice versa).  A small amount of retraining of the visible light model (using a few samples from a near-infrared dataset) yields a tuned network accurate in both settings. 
		
		For training and testing, this work uses the Casia v4 Interval, Notre Dame 0405, Ubiris v2, and IITD datasets.

\end{abstract}


\section{Introduction}
\label{sec:intro}

The extraction of consistent and identifiable features from the human iris is known as iris recognition.  
Iris recognition algorithms proceed in several phases: 1) segmenting pixels into iris and non-iris 2) extracting features from the iris that are likely to be stable 3) organizing features in a way that enables fast and accurate comparison.  We focus on the segmentation phase.  Errors in segmentation propagate to later stages of iris recognition~\cite{alonso2013quality}.  

Most segmentation algorithms work best with near-infrared (NIR) images~\cite{arsalan2018irisdensenet}\cite{liu2016accurate}. These algorithms can be used on visible light (RGB) images but their performance degrades~\cite{hofbauer2014ground}.  Segmentation algorithms also degrade in unconstrained environments where lighting and each individual's distance, orientation, and stability with respect to the imager vary.
Previous algorithms for working in unconstrained environments are dataset centric~\cite{proenca2010iris}.  \emph{The  goal of this work is to design iris segmentation that works in multiple environments.}
Prior approaches can be classified into three types: 
\begin{enumerate}[a)]
\item \emph{Specialized} approaches that use processing techniques developed for the iris.   Specialized approaches are fast and require no ground truth data. However, specialized algorithms require hand-crafted features or algorithms. These approaches have limited portability. 
 \item \emph{Hybrid} approaches that combine machine learning and specialized iris techniques to improved accuracy.  Hybrid approaches may require ground truth data.  
Hybrid algorithms attempt to enhance portability by pre-processing data to a standard representation before using a specialized approach.
 \item \emph{Learning} approaches that use general techniques for the entire process.  Learning based approaches rely on ground truth data to train classifiers.
Learning approaches often need a large amount of labeled training data. Previous approaches have been slow without a large accuracy improvement.  One advantage of learning approaches is the inheritance of general segmentation advances.
 \end{enumerate}

\noindent
No approach accurately segments irises in diverse environments.
We focus on learning based approaches and defer discussion of prior specialized and hybrid approaches to Appendix~\ref{sec:rel_work}.

An important learning based mechanism is the convolutional neural network or CNN which can be used for classification, feature extraction, and segmentation.  Hierarchical CNNs and Multi scale CNNs~\cite{liu2016accurate} fuse features from images at different scales. CNNs are natively translation invariant.  Arsalan et al. propose a two stage segmentation scheme  where coarse bounding box around the iris is extracted and used to train a CNN to classify iris pixels~\cite{arsalan2017deep}. 
Fully convolutional networks search along the entire image and calculate probabilities of pixels being iris or non-iris. A bounding box around the circular iris can reduce search space reducing complexity and increasing segmentation accuracy.



\paragraph{Our Approach} 
We propose a new iris segmentation scheme based on convolutional neural networks. Our approach is based on two key ideas:
\begin{enumerate}
\item The circular and continuous nature of the iris makes it an ideal candidate for bounding box localization. A recent CNN based segmentation scheme~\cite{arsalan2017deep} used this idea to detect an ROI (Region of interest) around the iris manually by using edge detection and binarization. Our approach handles this as a learning problem without using specific transforms. \item Bounding box fitting and classification have been treated as separate problems~\cite{girshick2014rich}. It is more intuitive to unify these problems so each each item can be solved in parallel increasing speed and reducing redundancy by sharing features. Mask inference, that is, deciding which bits are iris bits, can now be done more efficiently.  The search space has been reduced once the bounding box class has been predicted. Importantly, we use a  loss function that is a cumulative of the bounding box, classification and mask inference losses.
\end{enumerate}

To implement these ideas, we start with the recent Mask R-CNN framework~\cite{he2017mask}. He et al. proposed the Mask R-CNN framework for large scale segmentation involving numerous classes.  We find this framework can be adapted to iris segmentation. Importantly, the framework natively produces a per pixel segmentation, a \emph{mask}, ideal for iris mask extraction.

We demonstrate our techniques using four datasets: Casia v4 Interval~\cite{casia}, Ubiris~\cite{pro09a}, ND-0405~\cite{Bowyer2009TheNI}, and IITD~\cite{kumar2010comparison}.   We find that Casia v4 and IITD are similar in many aspects.  Ubiris is the most difficult and different dataset as it is collected in an unconstrained environment.  Jalilian et al. considered Casia v4, Casia v5 aging, and IITD.  We believe our dataset collection is more diverse; we directly compare with Jalilian et al. where possible.

\paragraph{Single dataset networks}
We first train and test our design on images from each dataset.
Our scheme achieves comparable segmentation accuracy as previous approaches when trained and tested on an individual dataset.  The main advantage of our approach is high accuracy when training and testing on different data sets.  As a comparison point, recently Jalilian et al.~\cite{jalilian2017domain} used \emph{domain adaption} to allow a trained CNN to work in multiple environments.  Their approach requires samples from each new environment.
Our accuracy is higher than Jalilian et al.~\cite{jalilian2017domain} without performing explicit domain adaption.

\paragraph{Towards a Universal Network}
However, our single network approach demonstrates low accuracy when training on an RGB dataset and testing on an NIR dataset (and vice versa).  To remedy this situation, we then develop a second model to segment irises from images across all four datasets used. This single model can segment images from any of the four datasets. We discuss the design of this network further in Section~\ref{sec:tuned model}.

\paragraph{Performance}
In addition to segmentation performance we briefly remark on the speed of our implementation.
Our implementation speed varies from 8 fps to 20 fps depending on the dataset used.  The speed is largely determined by the resolution of the iris image. While this speed is not sufficient to keep pace with video,  it is faster than previous learning methods and static methods. Liu et al.~\cite{liu2016accurate} report a inference time of 0.2 seconds for their CNN scheme while our implementation roughly segments in 0.09 seconds, both works using the Ubiris dataset. Training time can be fine tuned, dependent on the application requirements. We have datasets from different wavelengths and different camera sensors.     

\paragraph{Organization}
The rest of this work is organized as follows, Section~\ref{sec:design} describes our overall design (including our single dataset networks), Section~\ref{sec:evaluation} describes the evaluation methodology and results on single dataset trained model, Section~\ref{sec:tuned model} describes our tuned network and results, and Section~\ref{sec:conclusion} concludes.  Specialized and hybrid segmentation schemes are discussed in Appendix~\ref{sec:rel_work}.


\section{Design}
\label{sec:design}
\subsection{The Mask R-CNN Framework}

\begin{figure*}[!thp]
	\centering
	\includegraphics[scale=0.49]{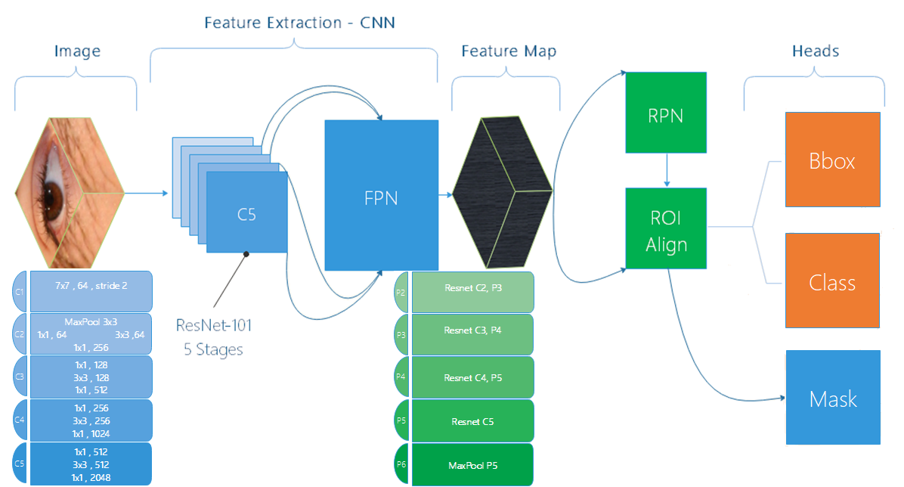}
	\caption{Simplified overview of the Mask R-CNN framework. Below the five stages of C5/ResNet and FPN are listed below the component.  In the FPN stages $C4$ refers to the fourth stage of ResNet while $P4$ refers to the fourth stage of FPN.  Note that in MR-CNN bounding box inference, classification, and mask creation are inferred simultaneously.}
	\label{cnnArch}
\end{figure*}


The starting point for our network is the recent Mask R-CNN framework~\cite{he2017mask}. The Mask R-CNN framework augments the region based convolutional neural network (R-CNN).  R-CNN takes a two step approach to segmentation.  It first finds regions of interest~(ROI) using selective search~\cite{girshick2014rich}.  Then it finds CNN features for these ROI which are classified using a support vector machine (SVM). The Fast R-CNN~\cite{girshick2015fast} introduced pooled ROIs while unifying the bounding box and classification heads. The Faster R-CNN~\cite{ren2015faster} speeds up the region proposal process by running a pass of a CNN to obtain feature maps. Proposals are generated from the feature maps by sliding fixed sized anchors to obtain regions with probable objects. A bounding box regressor fits bounding boxes over the regions. Classification of the bounding boxes is handled by a dense layer. MR-CNN adds a third head to the Faster R-CNN, it predicts segmentation masks.  

The full MR-CNN framework is summarized in Figure~\ref{cnnArch}.  
The MR-CNN framework performs feature extraction using two components: a feature extractor and a feature scaling component.  We use ResNet-101~\cite{he2016deep} for feature extraction and the FPN (Feature Pyramid Network) \cite{lin2017feature} for the scaling component.

The MR-CNN has some improvements over previous works. It uses a Resnet backbone for feature extraction. The ResNet has skip connections inside layers which adds previous activations to the current operation allowing for deeper networks. This allows the gradient to flow back into the network which would have otherwise diminished. The feature extractor can thus be trained for longer time periods. 

ResNet is a five stage CNN architecture.  Each stage has layers of convolution neurons.  Each stage produces activation maps at a different scale (decreasing by a factor of $2$).  The architecture outputs the result of the last four stages of ResNet.  The purpose of outputting multiple stages is to allow the five stage FPN to make the network more resilient to changes in scale. This is done by inferring at five different scales per iris image.

The output of the FPN serves as the feature map. A Region Proposal Network proposes regions in the form of ROIs by sliding anchors of fixed sizes over the feature maps. These ROIs are independently pooled to form fixed dimension  (7x7) feature maps. The ROI align layer makes the pooled representation of the feature maps aligned with the previous layers by using bilinear interpolation. Classification, bounding box prediction and mask inference is done using the pooled feature maps. A mean binary cross-entropy loss is used to calculate a mask per class so there is no competition among classes for mask inference. The class prediction uses a softmax loss function while the bounding box head uses a smooth L1 loss. 

We use the same Intersection over Union threshold and the FPN implementation as the Mask R-CNN paper~\cite{he2017mask}.

\subsection{Our Network Design}
In this subsection, we discuss the design of our single data set networks building on  the Mask R-CNN framework.

Since all four datasets contain iris images without other facial features such as the nose and lips there are fewer candidate bounding boxes accepted for training. As such the bounding box head trains fairly quickly.

  We use a pre-trained model for general semantic and instance segmentation for our work, a model trained~\cite{mrcnn} on the Microsoft COCO dataset~\cite{lin2014microsoft}. We retrain this model on our datasets.

Our pre-trained model has 80 classes. We expect the complex pre-trained  model to port quickly. 
Since our use case has one class to segment, starting from the pre-trained model should decrease training time.

MR-CNN needs bounding boxes per image, our ground truth has masks only. We create bounding boxes from each mask so that the box encompasses the entire mask with 2 pixels to spare on each side. 

Since we are performing binary classification (iris/not-iris), we can further reduce training time by reducing ROIs per image. We reduce ROIs by reducing anchors (the number of different shaped anchor strides that are performed across the image) and using fewer candidate ROIs.  Specially, we only use square anchors of 64, 128 and 256 pixels. This reduced number of candidate ROIs only affects the mask accuracy by $2\%$ while speeding up training time by $40\%$.

Inference (mask extraction) time is a crucial metric for CNN architectures.  Ideally, the architecture should infer at camera speed of 30 frames per second. The MR-CNN paper reports a mask inference speed of 5 frames per second. Our implementation of the Mask R-CNN segments an iris in 0.09 seconds for the Ubiris dataset which roughly equates to 11 frames per second. We attribute this speed improvement to having only one class to segment and lower resolution images.

Instance segmentation is an important aspect of the framework where segmented objects are categorized as instances of a single class. The MR-CNN is an instance-segmentation first framework. This could be useful to segmenting both iris instances from a single image. Iris instance segmentation helps with iris recognition.

For iris segmentation, the main difference between Mask R-CNN and an FCN is the bounding box head.  To test this distinction, we trained our models where the bounding was always set to the perimeter of the image.  This change degrades the F1 score.

Many iris processing techniques use data augmentation to help in the training of the network.  The goal of augmentation is avoiding over-fitting by adding variation to the training dataset. As an example, some schemes blur the image to contrast changes. Although this process adds a pre-processing step it further improves segmentation accuracy by adding additional samples to the training data. Our augmentation is simple, we randomly flip $50\%$ of iris images horizontally during training. We also perform a trivial transform on the grayscale NIR images.  We upscale them to RGB by copying the grayscale value in the R,G and B bins. This allows a single network for inference on different datasets.

\paragraph{\textbf{Datasets}}
Our work uses the CASIA, IITD, UBIRIS and the Notre Dame datasets. Since our approach uses supervised learning we restrict to the subset of this datasets where ground truth is available.  We utilize segmentation ground truth datasets created by Hofbauer et al.~\cite{hofbauer2014ground}

\textbf{CASIA Iris Image Database version 4~\cite{casia}}
This is an NIR (Near infrared) dataset which consists of 2655 iris images from 249 subjects. The images are of 320 by 280 pixels.  Images were collected in an indoor constrained environment using a close up infrared iris camera which leaves an easily distinguishable circular reflection on the pupil across all images due to a circular LED array. 

\textbf{IIT Delhi Iris Database version 1~\cite{kumar2010comparison}}
This dataset has NIR images of resolution 320 by 240 taken using a JIRIS JPC1000. Images are taken in a constrained environment with little variation.

\textbf{ND-IRIS-0405~\cite{Bowyer2009TheNI}}
The Notre Dame database contains NIR images of resolution 640 by 480 pixels. These images are also taken in a controlled environment with little variation. For the ND dataset we use a small subset of the original dataset where groundtruth is provided by Hofbauer et al.~\cite{hofbauer2014ground}.

\textbf{UBIRIS version 2~\cite{pro09a}}
The UBIRIS database has 11101 iris images from 50 different subjects with resolution of 400 by 300 pixels. This dataset has wide variation among images with off-angle, reflections and imaging distance among the varying parameters introducing some realness to the database. The images are also RGB and do not present a clear iris pattern as opposed to the other datasets.

\begin{table}[t]
	\small
	\centering
	\caption{Dataset Statistics.  Each dataset is restricted to images which have segmentation ground truth~\cite{hofbauer2014ground}.}
	\label{tab:datasets}
	\begin{tabular}{|l|c|c|c|r|}
		
		\hline
		\TspcBig   Dataset & Resolution  & NIR & Realistic & Images\\
		\hline
		\TspcBig Casia v4 Interval     & 320x280 & Y & N & 2655  \\
		\TspcBig Ubiris v2      & 400x300 & N & Y & 2250  \\
		\TspcBig NotreDame   & 640x480 & Y & N & 837  \\
		\TspcBig IITD  & 320x240 & Y & N & 2240  \\
		\hline
		
		\hline
		
	\end{tabular}
	
\end{table} 

\paragraph{Metrics}
We use F-Measure and E1 score from the NICE competition~\cite{nice} to evaluate segmentation accuracy (see Table~\ref{tab:eval_parameters}). The E1 score is the normalized XOR of the binary predicted mask and the ground truth. The E1 score measures what fraction of pixels disagree. The F-measure is the harmonic mean of precision and recall. The E1 rate is normalized by image dimensions making it incompatible between datasets with differing dimensions. We use the F1 score for cross dataset accuracy comparison.

\begin{table}[htb]
	\small
	\caption{Evaluation Metrics.  TP is true positive: the fraction of pixels that are correctly segment as iris.  TN is true negative: the fraction of pixels that are correctly segmented as non-iris.  FP and FN are false positive and false negative and are defined in as 1-TP and 1-TN respectively.  }
	\label{tab:eval_parameters}
	
	\centering
	{\begin{tabular}{|l| c |}
			\hline
			\textbf{\TspcBig Measure} & \textbf{Value}           \\
			\hline \hline
			\Tspc         Precision             &     TP/(TP+TN)        \\

			\hline
			\Tspc         Recall             &     TP/(TP+FP)        \\
			
			\hline
			\Tspc         F-Measure             &     $\dfrac{2*\text{Recall}*\text{Precision}}{\text{Precision}+\text{Recall}}$       \\
			\hline
			\Tspc         E-Nice1             &     $\dfrac{\#\text{ pixels that differ from ground truth}}{\text{Image Size}}$      \\
			\hline
	\end{tabular}}
	\vspace{-0.2cm}

\end{table}

\section{Single Dataset Network Results}

\label{sec:evaluation}
\label{sec:evaluation method}

In this section we describe our performance in training networks for each of the four datasets described above.  
We use a 3 step training methodology where 1) the heads, 2) stages 4 and up of the ResNet and 3) all stages are trained in sequence. All datasets are split into 40-30-30 disjoint sets for training, testing, and validation respectively.   Table~\ref{tab:single dataset} which lists performance of the individual datasets with respect to all metrics listed in Table~\ref{tab:eval_parameters}.   

Performance on individual datasets is competitive with state of the art.  Arsalan et al.~\cite{arsalan2018irisdensenet} report slightly better performance on Ubiris than our techniques.  We note that Arsalan et al. use ground truth from the NICE competition while we use ground truth provided by Hofbauer et al.~\cite{hofbauer2014ground}. 

\begin{table}[!th]
	
	\caption{Segmentation Rates four fold cross validation, trained on 40\% and Tested on 30\% of the same dataset.  Columns present the mean, $\mu$ and variance, $\sigma^2$, for each metric introduced in Table~\ref{tab:eval_parameters}.  E1 score should not be compared across datasets as it is scale dependent. The results from 1 fold and 4 fold validation don't differ.}
	\label{tab:single dataset}
	\centering
		\vspace{+0.2cm}
	\begin{tabular}{| l |r r |r r|r r |r r |}
		\hline
		& \multicolumn{2}{c|}{\textbf{Prec. \%}} & \multicolumn{2}{c|}{\textbf{Rec. \%}} & \multicolumn{2}{c|}{\textbf{F1 \%}}& \multicolumn{2}{c|}{\textbf{E1  \%}}\\
		\hline
		& {$\mu$}. & $\sigma^2$.  & $\mu$. & $\sigma^2$. & $\mu$. & $\sigma^2$. & $\mu$. & $\sigma^2$.\\
		\hline
		Casia       & 98.2 & 0.01 & 93.6 & 0.03 & 95.8 & 0.01 & 0.024 &0.003\\
		Ubiris      & 95.9  & 0.12 & 93.4 & 0.08 & 94.6 & 0.05 & 0.77 &0.004\\
		ND   & 97.9 & 0.02 & 92.7 & 0.04 & 95.2 & 0.02 & 0.16 &0.005\\
		IITD  & 98.5 & 0.01 & 94.1 & 0.03 & 96.2 & 0.01& 0.023 &0.002\\
		\hline
		
		\hline
		
	\end{tabular}
	\vspace{+0.2cm}

\end{table}

We run adaptability experiments using our four datasets and trained models. We generate a 4x4 grid by training on a single dataset and then testing on all four datasets.  Table~\ref{tab:cross performance} shows the F1 score (the harmonic mean of precision and recall) for each dataset combination.  Our system demonstrates portability across datasets. As a comparison, Jalilian et al. report an F1 score of .81 when training on Casia v4 Interval and testing on domain adapted IITD images (.71 without domain adaption).  Our system achieves an F1 score of $.95$ for the same pair.

This baseline system achieves adequate performance when moving between IITD and Casia v4 Interval.  However, training on just Ubiris (the most difficult and diverse dataset) achieves inadequate performance segmenting on other datasets.  Similarly, all models that are not trained on Ubiris demonstrate poor performance on Ubiris.  

Segmentation F1 scores of 85\% are achieved fairly quickly after training the heads for 2 epochs. After two epochs there is a depreciating accuracy return.


\begin{table}[t]
	\footnotesize
	\centering
	\caption{Segmentation Rates Trained and Tested on different datasets - F1 Score. The columns list the mean, $\mu$, and variance, $\sigma^2$, when the network trained on the dataset in the \emph{column} was tested on the dataset in the \emph{row}.  The diagonal represents testing and training on the same dataset. }
	\vspace{+0.2cm}
	\begin{tabular}{|l |rr|r r |r r |r r|}
		\hline
		\TspcBig \textbf{Dataset} & \multicolumn{2}{r|}{\textbf{Casia }} & \multicolumn{2}{r|}{\textbf{Ubiris }} & \multicolumn{2}{r|}{\textbf{ND }}& \multicolumn{2}{r|}{\textbf{IITD }}\\
		\hline
		\TspcBig  & {$\mu$}. & $\sigma^2$.  & $\mu$. & $\sigma^2$. & $\mu$. & $\sigma^2$. & $\mu$. & $\sigma^2$.\\
		\hline
		\TspcBig Casia      & 95.8 & 0.01 & 55.5 & 2.54 & 95.1 & 0.01 & 95.3 &0.04\\
		\TspcBig Ubiris   & 63.5  & 0.39 &94.6 & 0.05 & 81.0 & 2.25 & 70.4 &1.56\\
		\TspcBig ND   & 93.1 & 0.11 & 71.2 & 1.60 & 95.2 & 0.02 & 93.9 &0.05\\
		\TspcBig IITD  & 95.2 & 0.03 & 67.4 & 2.47 & 93.7 & 0.08 & 96.2 &0.01\\
		\hline
		
		\hline
	\end{tabular}
	\label{tab:cross performance}
\end{table}

\begin{table}
	\caption{Recent Ubiris segmentation rates. Bazrafkan et al.~\cite{bazrafkan2017end} infers on only 10\% of the dataset while training on 70\%, we find 10\% of the Ubiris dataset to be too low to show the unconstrained nature of the dataset. The works~\cite{arsalan2017deep,liu2016accurate,zhao2015accurate,arsalan2018irisdensenet} use the mask dataset from the Nice competition (we use ground truth from~\cite{hofbauer2014ground}).  They do not report F1 score. Our score is four fold cross-validation, trained on 60\% and tested on 20\%.}
	\label{tab:ubiris seg}
	\vspace{+0.05cm}
	\centering
	\begin{tabular}{|l |r|r|}

		\hline
		\TspcBig \textbf{Method} & \textbf {F1 Score} & \textbf {E1 score}\\
		\hline
		\TspcBig Ours & 94.8 &0.74\\
		\hline
		\TspcBig U-shaped CNN \cite{bazrafkan2017end} & 93.9 &0.70     \\
		\hline
		\TspcBig IrisDenseNet \cite{arsalan2018irisdensenet} & - &0.70     \\
		\hline
		\TspcBig CNN~\cite{arsalan2017deep}&-&0.82\\
		\hline
		\TspcBig MFCNN~\cite{liu2016accurate}&-&0.90\\
		\hline
		\TspcBig Total Variation~\cite{zhao2015accurate}&-&1.13
		\\
		\hline
		\TspcBig Conv Encoder-Decoder \cite{jalilian2017iris} & 86.3 & 1.87     \\
		\hline
		
	\end{tabular}
\end{table}

We compare our scheme to recent Ubiris segmentation schemes in Table~\ref{tab:ubiris seg}.  Our scheme has comparable performance on Ubiris with the benefit of cross dataset portability.
The results in Table~\ref{tab:single dataset} are from a model trained on 40\% of each dataset. Table~\ref{tab:ubirisNew} presents results from testing on a 60-20-20 train-test-validation split for comparison with prior work. The results for the Casia, IITD, Notre Dame dataset show no significant change when trained on 60\% of the dataset therefore we only show the results from the Ubiris dataset in Table~\ref{tab:ubirisNew}.

\begin{table}
	\caption{Ubiris model trained on 60\% of the dataset. Results are from four-fold cross-validation.}
	\label{tab:ubirisNew}
	
	\centering
	\begin{tabular}{| l |r r |r r|r r |r r |}
	\hline
	& \multicolumn{2}{c|}{\textbf{Prec. \%}} & \multicolumn{2}{c|}{\textbf{Rec. \%}} & \multicolumn{2}{c|}{\textbf{F1 \%}}& \multicolumn{2}{c|}{\textbf{E1  \%}}\\
	\hline
	& {$\mu$}. & $\sigma^2$.  & $\mu$. & $\sigma^2$. & $\mu$. & $\sigma^2$. & $\mu$. & $\sigma^2$.\\
	\hline
	Ubiris      & 96.2  & 0.12 & 93.5 & 0.08 & 94.8 & 0.04 & 0.74 &0.004\\
	
	\hline
	
	\hline
	
\end{tabular}
\end{table}

\section{Universal Model}
\label{sec:tuned model}
In the last section we showed our networks trained on a single dataset demonstrated portability especially between NIR datasets.  In this section, we show how to tune a model that segments well on all four datasets.

 There is an affinity between the CASIA, IITD and Notre Dame datasets with all cross-dataset F1 scores between them being at least $90\%$. The Ubiris and the Notre Dame dataset have similar \emph{scale} the fraction of the image occupied by the iris.  We posit this is the reason by higher F1 scores on the ND-Ubiris pairs (compared to other pairs involving Ubiris). 
 
Ubiris here is the most difficult dataset and its models' capability does not carry over to the the CASIA and IITD datasets.  We believe the two most relevant factors for model portability are the scale of the iris and the ``hardness'' of a dataset. Two factors that contribute to hardness is whether a dataset is NIR or  RGB and whether is environmentally constrained. 

The Casia and IITD datasets expose the iris wholly and the majority of the image holds the iris relative to the Ubiris and the Notre Dame datasets where the images are from a distance and the iris covers a smaller area. This is an apparent and intuitive problem with any CNN architecture.  The FPN component is designed to handle this problem and make the network scale invariant.  Scale invariance is further enhanced by training on both Ubiris and Casia.

We choose to train a Ubiris-Casia model to generalize towards all the datasets. Ubiris was chosen as the base dataset due to its hardness and Casia used for retraining due to its NIR nature. Ubiris and Casia also differ in scale by 20.7\%. This difference will help in making the model more scale invariant.  Additionally, this presents the model with both NIR and RGB features.

We use a 40-30-30 train-test-validation split of the Ubiris dataset to train the generic model. We then use 100 images from the Casia dataset to retrain the model.  The effect is to tune it towards the NIR datasets.
(We have also provided results from testing on a 60-20-20 train-test-validation split for comparison with prior work.)

\begin{figure*}[t]
	\centering
	\includegraphics[scale=0.99]{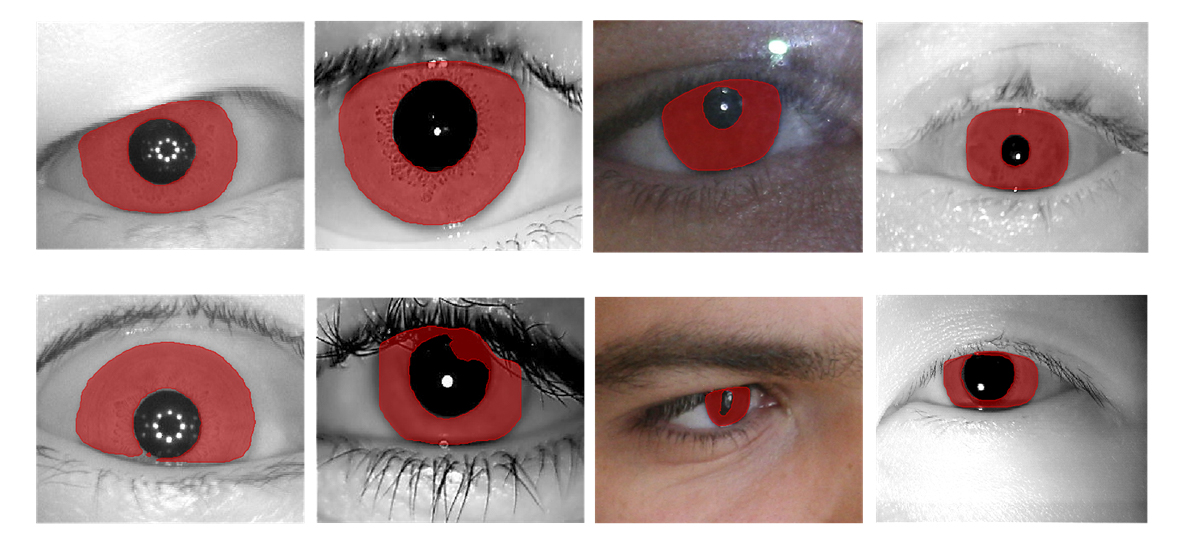}
	\caption{Segmented Iris Masks from the Casia, IITD, Ubiris and Notre Dame datasets respectively. The final tuned Ubiris-Casia model was used here. The top row shows the Average-Best case masks while the bottom row shows the worst case masks.}
	\label{fig:example segmentations}
\end{figure*}

We first train the COCO model on the Ubiris training set using the same three stage training regime described Section~\ref{sec:design}.  This consists of training the 1) the heads, 2) stages 4 and up of the ResNet and 3) all stages are trained in sequence.  We then retrain the heads of the model again on the Casia training set for 2 epochs. We use $100$ images from the Casia dataset for tuning.
\begin{table}[t]
	\small
	\centering
	\caption{Model trained on Ubiris with tuning on Casia.  This tuned model was then tested on all four datasets.  Results are the mean, $\mu$, and variance, $\sigma^2$ of F1 Score. Results are from four-fold cross-validation.  The second row of the table lists performance of the untuned model as reported in Table~\ref{tab:cross performance}.}
	\label{tab:tuned model}
	
	\begin{tabular}{|l |rr|r r |r r |r r|}
		\hline
		\TspcBig \textbf{Dataset} & \multicolumn{2}{r|}{\textbf{Casia }} & \multicolumn{2}{r|}{\textbf{Ubiris }} & \multicolumn{2}{r|}{\textbf{ND }}& \multicolumn{2}{r|}{\textbf{IITD }}\\
		\hline
		\TspcBig  & {$\mu$}. & $\sigma^2$.  & $\mu$. & $\sigma^2$. & $\mu$. & $\sigma^2$. & $\mu$. & $\sigma^2$.\\
		\hline
		\TspcBig Ubiris  Tuned  & 95.9  & 0.01 & 93.2 & 0.17 & 95.2 & 0.03 & 95.6 &0.04\\
		\TspcBig Ubiris  Untuned  & 63.5  & 0.39 &94.6 & 0.05 & 81.0 & 2.25 & 70.4 &1.56\\
		\hline
		
		\hline
	\end{tabular}
%
\end{table}
Figure~\ref{fig:example segmentations} shows examples of where the tuned model performs well on each dataset (top row) and poorly on each dataset (bottom row).  
This tuned model sees a slight decrease in the Ubiris F1 score (from $94.6\%$ to $93.2\%$).  However, as shown in Table~\ref{tab:tuned model}, Casia and IITD datasets see a drastic increase in F1 (at least $95\%$). The model is accurate on both IITD and ND datasets without being trained on samples from these datasets.
\subsection{Discussion}

The adaption of the network to the four datasets can be attributed to the FPN and the pixel intensities. The FPN deals with the difference in scales. The Resnet activations are tuned towards the circular nature of the iris and the pupil and are generic towards the four datasets. The NIR images are converted to RGB (for consistency) by pasting the grayscale value in the RGB bins, thus re-training the network on a few samples from one NIR dataset will expose the network to the differing intensity values and will be tuned to segment NIR images. Making the image corners as ground truth for bounding boxes degrades the segmentation accuracy by 3\% on average across the datasets. The mask head mimics an FCN except with a different loss function. 

\begin{figure}
	\centering
	\includegraphics[scale=0.35]{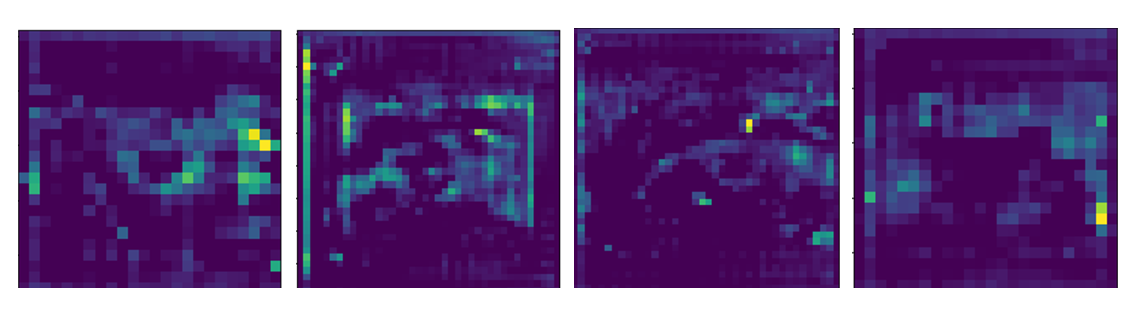}
	\caption{Activation maps from the fourth layer of the ResNet. The activations show the circular shape of the iris for all four datasets. IITD, Casia, ND and Ubiris in order.}
	\label{fig:example activations}
\end{figure}

 We find that we can group datasets based on the scale of the iris. Ubiris has the smallest scale at 6.98\% of the image on average. IITD has the largest scale at 30.68\%. Notre Dame has 9.4\% following closely to Ubiris while the Casia dataset has the scale at 27.7\%. It is important to note that Ubiris also has a high variance in its scale. Casia and IITD both being NIR datasets with similar scale are similar F1 scores. They also show similar F1 scores if Casia is used to infer IITD images and vice-versa. ND also follows suit but is less than optimal due to a smaller scale of the iris.
 
From the Table ~\ref{tab:cross performance} we see that models trained on IITD, ND and Casia can be used to infer on each other. This can primarily be attributed to the pixel intensity values being close on average and the feature extraction backbone being trained to segment circular irises. An example activation map for the four datasets from the fourth Resnet layer can be seen in Figure ~\ref{fig:example activations}. They all show circular activations. The different sized irises between datasets are handled by the FPN with its different scaled outputs. The portability of the model can also be attributed to the mask head being decoupled from the bounding box and classification. This is different from an FCN where class and mask prediction is coupled resulting in segmentation inaccuracy.

\begin{figure}[t]
\centering
	\includegraphics[scale=1.2]{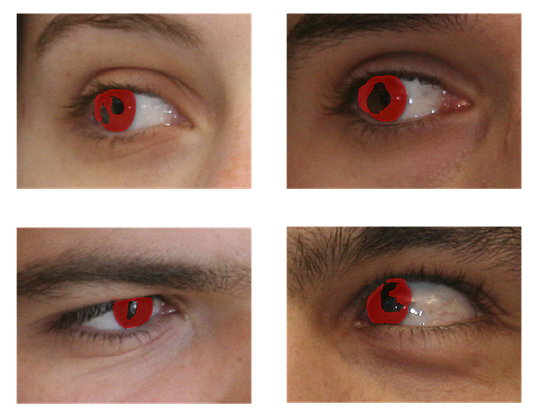}
	\caption{Some examples of the worst case segmentations. Images have F1 Scores below 80\%. We define a worst case segmentation as the 5th percentile in the CDF of F1 scores and the average-best case to be F1 scores above the 90th percentile. }
	\label{fig:ubiris bad segment}
\end{figure}

\begin{figure*}[t]
\centering
\begin{tabular}{c  c} 
\begin{minipage}{.48\textwidth}
  \centering
  \includegraphics[width=1.10\linewidth]{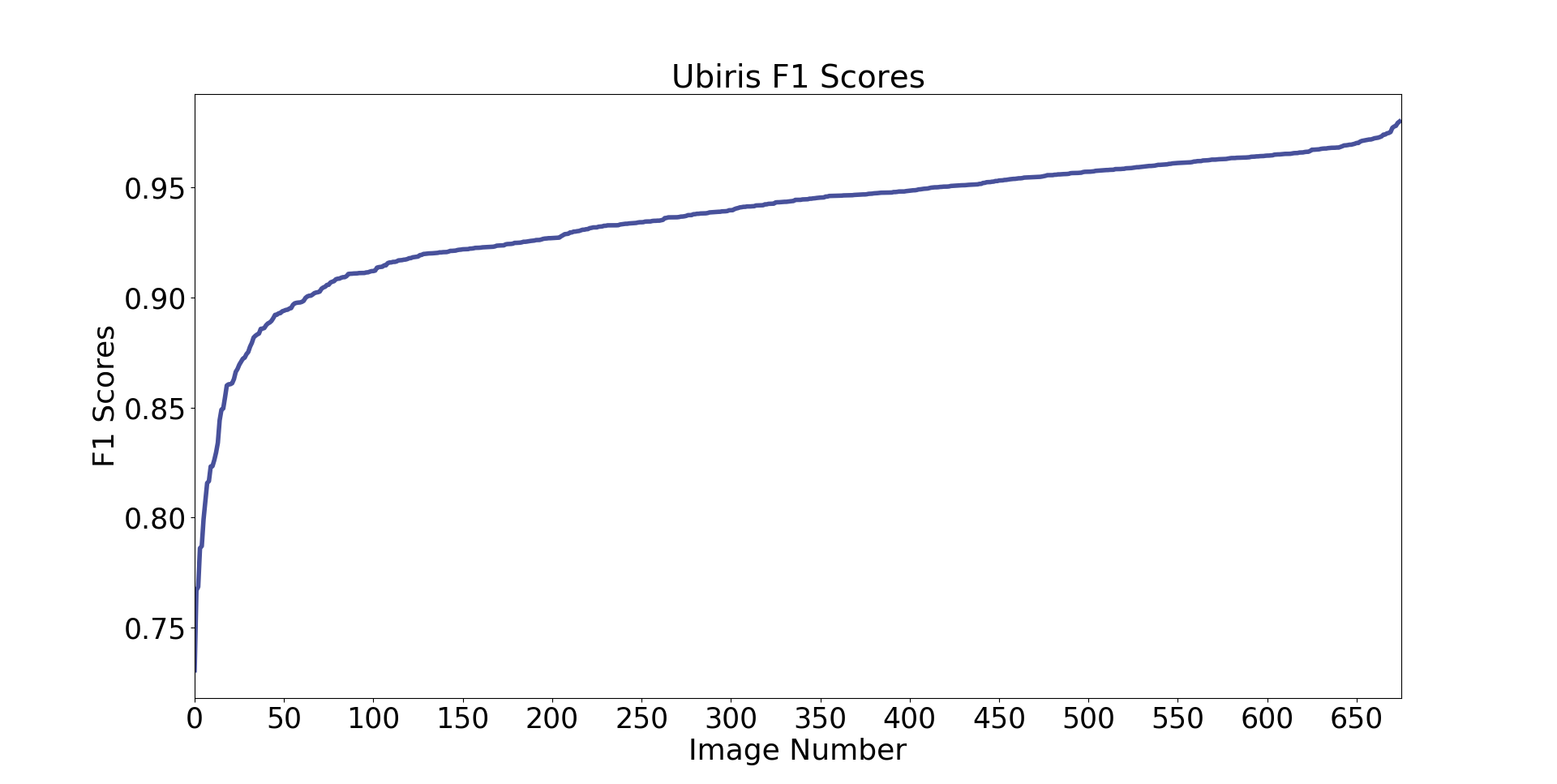}
  \label{fig:test1}
\end{minipage} & 
\begin{minipage}{.48\textwidth}
  \centering
  \includegraphics[width=1.1\textwidth]{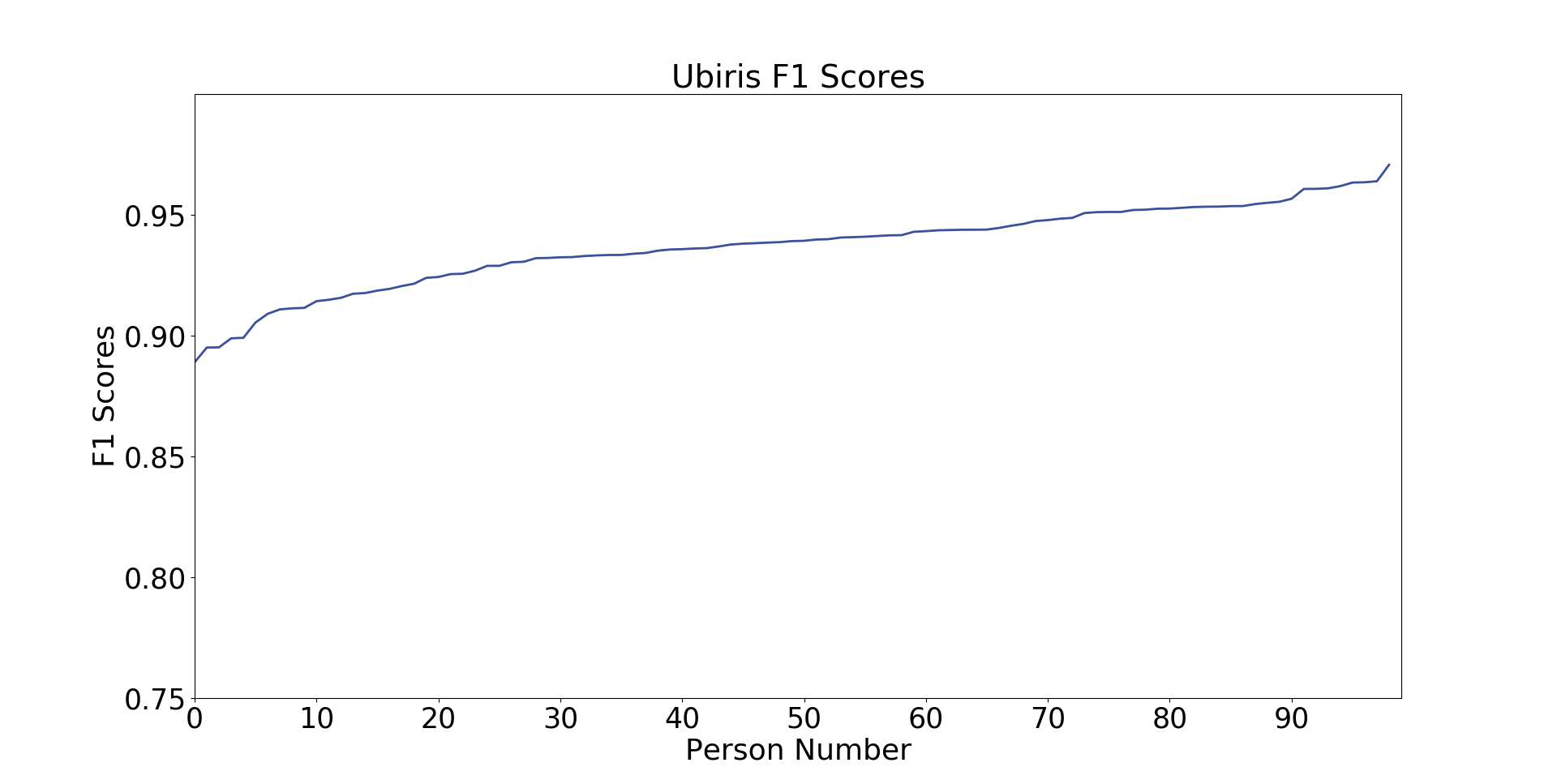}
  \vspace{.02in}
\end{minipage}
\end{tabular}
\caption{Cumulative distribution functions of F1 score for Ubiris dataset.  Left is a per image distribution function. The right is a per person distribution function.  Variance is smaller per person indicating that our network fails to segment particular images rather than performing poorly on individuals. Model trained on just Ubiris is presented here.}
  \label{fig:test2}
\end{figure*}

\begin{figure*}[h]
	\centering
	\includegraphics[scale=0.35]{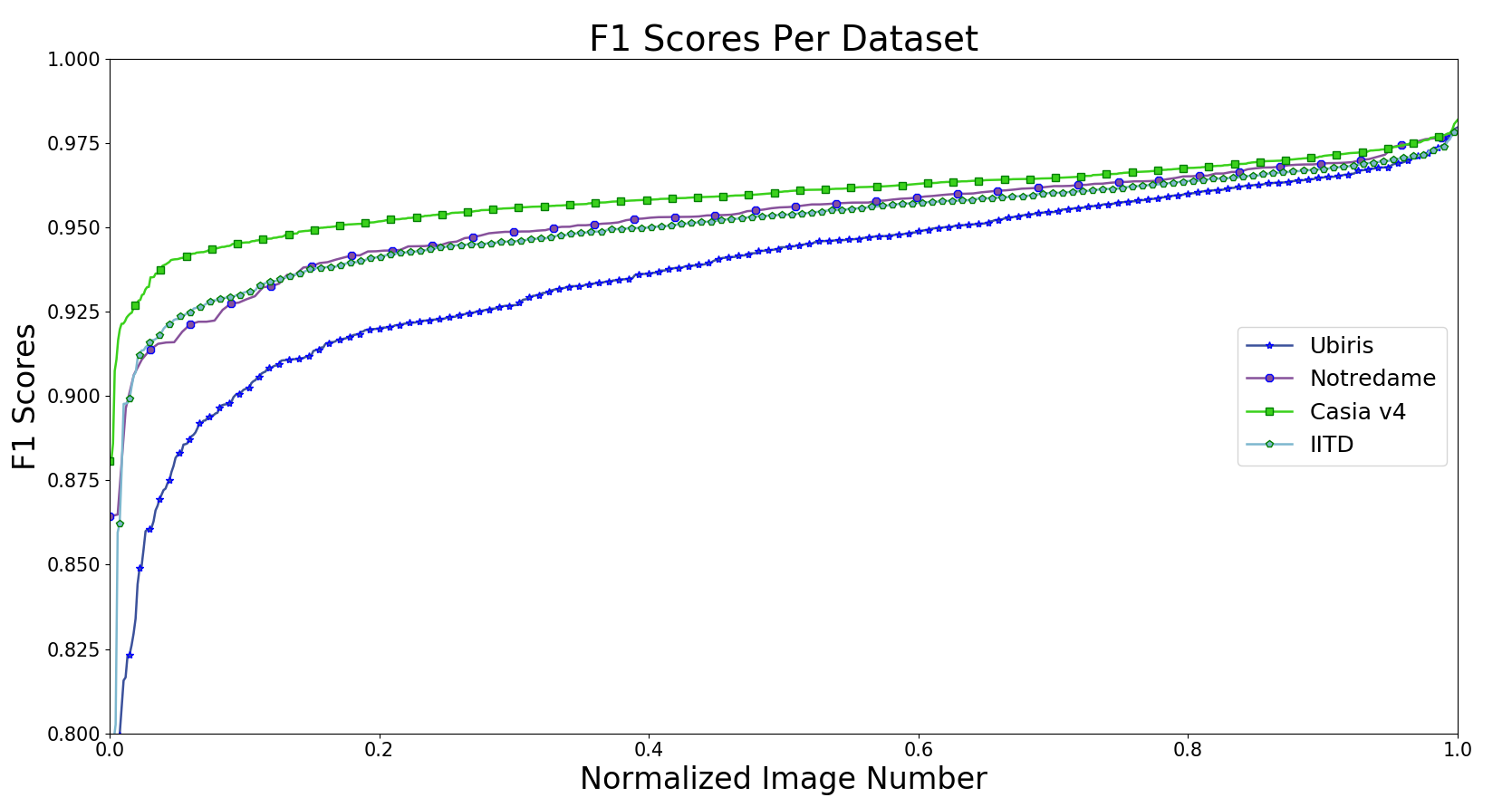}
	\caption{Sorted F1 scores from all four datasets. The distributions are based on the tuned model described in Section~\ref{sec:tuned model}.  Note that our system performs worst on Ubiris. }
	\label{fig:joint cdf}
\end{figure*}

The distribution of F1 scores for all four datasets on the tuned model is shown in Figure~\ref{fig:joint cdf}.  Since Ubiris is the worst performing dataset, we separate out the mean F1 score per person for Ubiris in Figure~\ref{fig:test2}.  Furthermore, we present additional instances of poor Ubiris segmentation in Figure~\ref{fig:ubiris bad segment}. This demonstrates that poor segmentation on Ubiris is a per image phenomena not a per person phenomena.

\paragraph{Inference speed}
The speed of the tuned model varies between 8-20 fps. The primary factor in determining the speed is the resolution of the image, Notre Dame takes 120 milliseconds seconds to segment. Ubiris follows closely at 90ms while Casia and IITD take 54ms. All inferences run on a laptop with an i7-7700HQ, 32GB DDR4 RAM, an NVidia GTX 1070. The IrisDenseNet~\cite{arsalan2018irisdensenet} does not report its inference time but their work is based on the SegNet~\cite{badrinarayanan2015segnet} which reports an forward pass time of 422ms with an image of dimension 360x480.

%

%
%
%


\section{Conclusion}
\label{sec:conclusion}

This work presents a new CNN based iris segmentation algorithm whose performance is close to best in class on representative datasets.  Our algorithm natively can be used to perform segmentation on a different dataset than was used for training.  Datasets where the iris is of similar scale demonstrate high accuracy in cross dataset experiments.  To create a ``universal'' algorithm we need to correct for differing scale of the iris within dataset.
  Our tuned algorithm primarily trains on Ubiris with a small amount of retraining on Casia.  This tuned network is accurate on all four datasets.  
  
  In addition to accuracy, our network is fast, segmenting up to $20$ frames per second.  Our result falls short of a video frame rate but is better than previous learning implementations. 
  
  As a \emph{future direction} we suggest using a similar CNN network to perform feature extraction to create a unified learning based iris recognition workflow. An \emph{open question} is whether we can domain adapt between RGB and NIR datasets with fewer samples or global dataset features. The entire segmentation weights could be recycled for iris recognition while simultaneously extracting features improving speed and decreasing redundancy.


	\section*{Acknowledgement}
	The authors would like to thank Comcast Inc. and Synchrony Financial for partial support of this research. The authors would like to thank the reviewers for their constructive feedback.
	\appendix
 \section{Additional Related Work}
 \label{sec:rel_work}

As discussed in the introduction, segmentation algorithms can be classified into specialized, hybrid, and learning.  We provide an overview of specialized and hybrid approaches below.

\paragraph{Specialized approaches} 
The first generation of segmentation algorithms assumed that the iris and pupil are circular.  Under this assumption circle/ellipse fitting techniques are used to find iris and pupil boundaries. Daugman's seminal work uses an integro-differential operator for iris segmentation~\cite{daugman2009iris}, in essence it tries to fit a circle over an iris exhaustively. Subsequent algorithms are based on integro-differential operators and the Hough transform~\cite{hough1962method,duda1972use,illingworth1988survey}.  Circle fitting via the Hough transform involves different regions voting for the best circle (which can be seen as an exhaustive search over a possible circle) \cite{wildes1997iris}.  New segmentation methods use different techniques to find candidate circles (for example, see Tan and Kumar~\cite{tan2013towards}). 

Methods that assume the iris and pupil are elliptical perform well in constrained environments where the individual faces the camera and actively participates in image collection.  However, these methods perform poorly in unconstrained environments.  As an example, in blurred iris images these algorithms fixate or diverge harshly, converging to edges besides the iris boundary.  Substantial pre-processing and post-processing is necessary to use these methods in unconstrained environments. 

Improvements to ellipse fitting have come in the form of generalizes structure tensors or GST~\cite{alonso2012iris}.  GSTs find an iris specific complex circular pattern and convolve this pattern with the collected image to find the iris and pupil. The GST method allows the discovered region to only approximate a circle. Zhao and Kumar first use a total variation model to regularize local variations~\cite{zhao2015accurate} and then feed this image into circular Hough transform.  Interestingly, Zhao and Kumar process the lower half and the upper half of the iris separately.  

Active contours~\cite{kass1988snakes} are used for general segmentation tasks. Instead of circle fitting, active contours find a high gradient between two sections in an image to indicate a boundary. Abdullah et al. show that with some iris specific modifications, active contours segment well~\cite{abdullah2017robust}.  While active contours do not assume the collected iris is circular they can still fixate on reflections and occlusions.


\paragraph{Hybrid approaches}
Machine learning techniques including neural networks have penetrated into fields that used specialized approaches.  
Many works segment the iris using a mix of general learning techniques and specialized iris techniques. The common approach is to use learning algorithms to do rough segmentation and post-process via specialized techniques to get the final segmented iris. 

Proenca performed coarse iris segmentation using a neural network classifier and refined this segmentation via polynomial fitting~\cite{proenca2010iris}. Radman et al. use a HoG (Histogram of Gradients) as input features to train a support vector machine (SVM)~\cite{radman2017automated}.  This trained SVM is then used on new images to localize the iris. Subsequent iris segmentation is done by employing the Growcut algorithm \cite{vezhnevets2005growcut} which labels pixels based on an initial guess. Adaboost~\cite{freund1997decision} based eye detection is common for identifying the iris region inside a facial image, Joeng et al.~\cite{jeong2010new} refined this technique by adding eyelid and occlusion removal algorithms. Tan and Kumar use Zernike moments as input features to train a neural network and SVM for coarse iris pixel classification~\cite{tan2012unified}.

\comments{
Most of these algorithms rely on specific features to be used to train a classifier for pixel classification or localizing the iris. Note that 
Domain adaption for reducing manually labeled training data is proposed by, a Fully connected CNN and adapted images are used for training purposes, this work shows that by adapting a previously unseen dataset on an existing labeled dataset acceptable performance can be achieved on the new dataset . }

	\bibliographystyle{splncs04}
	\bibliography{refs}

\end{document}